\documentclass[11pt]{article}

\usepackage{coling2020}
\usepackage{times}
\usepackage{latexsym}
\usepackage{booktabs}

\colingfinalcopy 

\usepackage[T1]{fontenc}
\usepackage{type1cm}
\usepackage[american]{babel}
\usepackage{url}
\usepackage{amssymb}
\usepackage{amsmath}
\usepackage{multirow}
\usepackage{xspace}
\DeclareMathAlphabet{\mathcalbf}{OMS}{pzc}{b}{n}
\usepackage{microtype}
\usepackage[utf8]{inputenc}
\usepackage{enumitem}

\usepackage[acronym]{glossaries}

\RequirePackage{color}
\definecolor{darkgray}{gray}{0.40}
\definecolor{mediumgray}{gray}{0.60}
\definecolor{lightgray}{gray}{0.95}
\definecolor{ultralightgray}{gray}{0.98}
\newcommand{\gr}{\color{mediumgray}}

\usepackage{graphicx}
\DeclareGraphicsExtensions{.pdf,.ai,.jpg,.png}
\setkeys{Gin}{pagebox=artbox}
\graphicspath{{./figures/}}

\newcommand{\hwfigure}[3][t!]{%
	\begin{figure*}[#1]
		\centering
		\includegraphics[scale=1.0]{#2}
    		\caption{#3}\label{#2}%
  	\end{figure*}
}

\RequirePackage{type1cm}
\RequirePackage{color}
\RequirePackage{soul}
\setstcolor{blue}
\definecolor{violet}{rgb}{0.5,0.0,0.5}

\newsavebox\bscombox
\newcommand{\bscom}[3][]{%
  \sbox{\bscombox}{\fontsize{8}{9}\selectfont#1#2#3}
  \noindent
  \st{#2}{\selectfont
    \color{blue}#3\ifx\\#1\\\else{\fontsize{8}{9}\selectfont\color{violet}[#1]}\fi
    }
  }

\makeglossaries
\newacronym{ca}{CA}{Computational Argumentation}
\newacronym{nlp}{NLP}{Natural Language Processing}

\newcommand{\weat}{\textsc{Weat}}

\begin{document}
%
%
\blfootnote{
    %
    %
    %
    %
    \hspace{-0.65cm}  
    This work is licensed under a Creative Commons
    Attribution 4.0 International License.
    License details:
    \url{http://creativecommons.org/licenses/by/4.0/}.
}

    \title{Argument from Old Man's View: Assessing Social Bias in Argumentation}

\author{Maximilian Splieth\"over \\
    Department of Computer Science \\
    Paderborn University \\
    Paderborn, Germany \\
    {\tt mspl@mail.upb.de} \\\And
    Henning Wachsmuth \\
    Department of Computer Science \\
    Paderborn University \\
    Paderborn, Germany \\
    {\tt henningw@upb.de} \\}

\date{}

\maketitle

\begin{abstract}
Social bias in language --- towards genders, ethnicities, ages, and other social groups --- poses a problem with ethical impact for many NLP applications. Recent research has shown that machine learning models trained on respective data may not only adopt, but even amplify the bias. So far, however, little attention has been paid to bias in computational argumentation. In this paper, we study the existence of social biases in large English debate portals. In particular, we train word embedding models on portal-specific corpora and systematically evaluate their bias using \weat, an existing metric to measure bias in word embeddings. In a  word co-occurrence analysis, we then investigate causes of bias. The results suggest that all tested debate corpora contain unbalanced and biased data, mostly in favor of male people with European-American names. Our empirical insights contribute towards an understanding of bias in argumentative data sources.
\end{abstract}

    \section{Introduction}

Social bias can be understood as implicit or explicit prejudices against, as well as unequal treatment or discrimination of, certain social groups in society~\cite{sweeney_transparent_2019,papakyriakopoulos_bias_2020}. A social group might be described by physical attributes of its members, such as sex and skin color, but also by more abstract categories, such as culture, heritage, gender identity, and religion. A typical, probably in itself biased, example of social bias is the old man's belief in classic gender stereotypes. In most cases, social bias is deemed negative and undesirable.

Recent research shows that bias towards social groups is also present in Machine Learning and \gls{nlp} models \cite{chang_bias_2019}, manifesting in the encoded states of a language model \cite{brown_language_2020} or simply causing worse performance for underrepresented classes \cite{sun_mitigating_2019}. Such bias has been studied for different \gls{nlp} contexts, including coreference resolution \cite{rudinger_gender_2018}, machine translation \cite{vanmassenhove_getting_2018}, and the training of word embedding models \cite{bolukbasi_man_2016}. In contrast, \gls{ca} has, to our knowledge, not seen any research in this direction so far. Given that major envisioned applications of \gls{ca} include the enhancement of human debating \cite{lawrence:2017} and the support of self-determined opinion formation \cite{wachsmuth:2017e}, we argue that studying social bias is particularly critical for \gls{ca}.

In general, social bias may affect diverse stages of \gls{ca}: In argument acquisition, for example, researchers may introduce social bias unintentionally, for instance, by collecting arguments from web sources that are only popular in a certain part of the world. This is known as {\em sample bias}~\cite{chang_bias_2019}. In argument quality assessment, a machine learning model may develop a {\em prejudicial bias} and judge arguments made by a certain social group better, for instance, because it considers features inadequate for the task, such as the gender~\cite{jones_machine_2019}. And in argument generation, a model might produce arguments that have an {\em implicit bias} towards a certain social group, for instance, because the features chosen are based on prior experience of the researchers and may not properly represent the whole population~\cite{fiske_social_2004}. As the examples indicate, social bias can, among other reasons, be caused by the source data and how it is being processed. As a starting point, this paper therefore focuses on social bias in the source data underlying \gls{ca} methods. In particular, we ask the following questions:
\begin{enumerate}
    \setlength{\itemsep}{0pt}
    \item
    What, if any, types of social bias are present in existing argument sources and how do different sources compare to each other in this regard?
    \item
    How much do certain groups of users contribute to the overall social bias of a source?
    \item
    What kinds of linguistic utterances contribute towards certain types of social bias?
\end{enumerate}
\noindent
Applications such as those outlined above usually rely on web arguments for scaling reasons. To study the questions, we hence resort to five English debate portals that give access to arguments on versatile topics: {\em 4forums.com}, {\em convinceme.net}, {\em createdebate.com}, {\em debate.org}, and {\em ChangeMyView}. All five have been deployed in \gls{ca} corpora \cite{abbott_internet_2016,durmus_corpus_2019,al_khatib_exploiting_2020}.

First, we analyze the general presence of social bias in each of the five debate portals. To this end, we train three custom word embedding models, one for each available corpus. Next, we evaluate the models for social bias using a widely used bias metric, called \weat~\cite{caliskan_semantics_2017}, and compare the results. We then inspect the debate.org portal more closely with regard to specific social groups. In particular, we group the texts based on the provided user information and apply the same evaluation.
Lastly, to gain a better understanding of what makes some texts more biased than others, we explore their language by analyzing word co-occurrences with group identity words in the texts.

Our findings suggest that all three corpora are generally biased towards male (compared to female) and European-American (compared to African-American) people. This bias is not only reflected in the \weat\ results, but also in unbalanced occurrences of identity words for certain social groups. More generally, we observe that the use of names as identity terms for social groups has unpredictable effects. With those insights, we contribute an initial understanding of social bias in sources of dialectical argumentative texts.
    \section{Related Work}

In recent years, research on different types of bias in natural language has received a considerable amount of attention. Media bias is one prominent example \cite{fan:2019}, particular the political bias of news articles \cite{chen:2020}. In various sub-fields of NLP, studies on media bias are concerned with analyzing techniques utilized by media outlets when reporting news. These include the framing of an event by phrasing the report with positive or negative terms, and selective reporting by including or omitting facts depending on the tone and (political) stance a media outlet wants to convey \cite{chen_learning_2018,hamborg_media_2020,lim_annotating_2020}. We do not target media bias here but {\em social} bias.

Social bias can emerge from pre-existing stereotypes towards any social group~\cite{sweeney_transparent_2019}, often leading to prejudices and discrimination. Stereotypes are understood as ``beliefs about the characteristics of group members''~\cite{fiske_social_2004}. They may be so powerful that they do not only limit the freedom of individuals, but also cause hate, exclusion, and --- in the worst case --- extermination~\cite{fiske_controlling_1993}. Even if stereotypes, and with that social biases, are individually controllable, they persist to this day; prominent examples of social groups that have historically been subject to bias are ethnicities, genders, and age groups~\cite{fiske_stereotyping_1998}. A major factor in carrying and reinforcing those biases is language~\cite{sap_social_2020}. In spoken and written argumentation and debates, biased language can be present if, for example, one sides argues in self-interest~\cite{zenker_experts_2011} to favor a certain social group or uses unbalanced arguments~\cite{kienpointner_problem_1997}.

As \gls{ca} methods are receiving more and more attention \cite{stede:2018}, it is important to understand how existing social biases influence them. One possible source is the human-generated data on which automated systems are trained and evaluated~\cite{chang_bias_2019}. In \gls{ca}, one of the main sources of data are online debate portals, both for research and for applications \cite{ajjour:2019a}.

Prior work has evaluated different properties of dialogical argumentation on debate portals. For example, \newcite{durmus_corpus_2019} evaluated the success rate of users in debates on debate.org based on prior experience, the users' social network in the portal, and linguistic features of their arguments. The authors find that information on a user is more informative in predicting the success compared to linguistic features. \newcite{al_khatib_exploiting_2020} retrieved all debates and posts from {\em Reddit's} discussion forum ChangeMyView to analyze the characteristics of debaters there. With this information, they were able to enhance existing approaches to predict the persuasiveness of an argument and a debater's resistance to be persuaded.

While the evaluation of debates notably misses work on social biases, other forms of bias have been studied. \newcite{stab_recognizing_2016}, for example, attempted to build a classifier that predicts the presence or absence of myside bias in monological texts. In contrast, this work offers an initial evaluation of social bias in dialogical argumentation by analyzing the posts of debate portals.

More generally, many approaches have been proposed to identify different types of social bias in word embedding models. In one of the first studies, \newcite{bolukbasi_man_2016} showed that pre-trained models contain gender bias. They found that it is revealed when generating analogies for a given set of words. This suggests that the distance between word vectors can act as a proxy to identify biases held by a model. Building on that notion, methods to automatically quantify the bias in a pre-trained word embedding model were presented. \newcite{caliskan_semantics_2017} introduce a metric named the {\em Word Embedding Association Test} (\weat) that adapts the idea of the {\em Implicit Association Test}~\cite{greenwald_measuring_1998}. Other methods include the {\em Mean Average Cosine Similarity (MAC)} test~\cite{manzini_black_2019}, the {\em Relational Inner Product Association (RIPA)} test~\cite{ethayarajh_understanding_2019}, the {\em Embedding Coherence Test (ECT)}, and the {\em Embedding Quality Test (EQT)}~\cite{dev_attenuating_2019}. For a more detailed literature review on detecting and mitigating biases in word embedding models, see \newcite{sun_mitigating_2019}. Our approach builds on the \weat\ metric to evaluate social biases in embedding models generated from debate portal texts.

Probably closest to our work is the study of~\newcite{rios_quantifying_2020}. Building on a method developed by \newcite{garg_word_2018}, the authors analyzed scientific abstracts of biomedical studies published in a time span of 60 years to quantify gender bias in the field and to track changes over time. For this purpose, they generated separate word embedding models for each decade in their data and evaluated them using the \weat\ metric. Using the RIPA test in addition, the authors identified the ``most biased words''~\cite{rios_quantifying_2020} of each model. While we will apply a similar method to detect social bias in textual data, our study differs in two main regards: First, instead of biomedical abstracts, we evaluate dialogical argumentative structures extracted from online debate portals and compare them to each other. Second, we additionally conduct a word co-occurrence analysis of the textual data as an attempt to get more insights into the \weat\ results; something that is notably missing in previous studies.
    \section{Data}
\label{sec:data}

To study social bias in argumentative language, we consider the dialogical argumentation found on online debate portals. As our analysis below is based on training word embedding models, which need sufficient data to find statistically meaningful co-occurrences, we resort to the three previously published corpora described below. The texts in these corpora have a similar argumentative structure, benefiting comparability:

\paragraph{IAC}

The {\em Internet Argument Corpus v2}~\cite{abbott_internet_2016} contains around 16k debates from multiple debate portals, including {\em createdebate.com}, {\em convinceme.net}, and {\em 4forums.com}. The debates tackle various topics that are led by users of the respective portals. As the number of arguments from the single portals are rather small compared to the other two corpora, we only consider this corpus as a whole.

\paragraph{CMV}

The {\em Webis-CMV-20} corpus~\cite{al_khatib_exploiting_2020} covers posts and comments of roughly 65k debates from the internet platform {\em reddit.com}, specifically from its subreddit {\em ChangeMyView}. Each debate consists of multiple comments and arguments, including the opening post in which a user states an opinion and arguments on a given issue and prompts other users to provide opposing arguments. The most convincing counter arguments can then be awarded by the initiator of the debate.

\paragraph{debate.org}

The corpus of \newcite{durmus_corpus_2019} is based on {\em debate.org}. It contains around 78k debates, each consisting of multiple arguments, written by over 45k users in total. In addition to the debates, the corpus has detailed self-provided user information, such as gender, ethnicity, and birth date. Adding information to one's profile is voluntary and, thus, neither available for all users in the corpus nor fully reliable. Still, we will use the available information to analyze arguments of certain user groups.
    \section{Experiments}
\label{sec:experiments}

This section presents the experiments that we carried out on the given data in light of our three research questions, as well as the underlying methodology. We describe how we train embedding models, evaluate their bias, and analyze the word co-occurrences causing the observed bias. Figure~\ref{process} illustrates the process.%
\footnote{The code for reproducing the experiments can be found at \url{https://github.com/webis-de/argmining20-social-bias-argumentation}.}

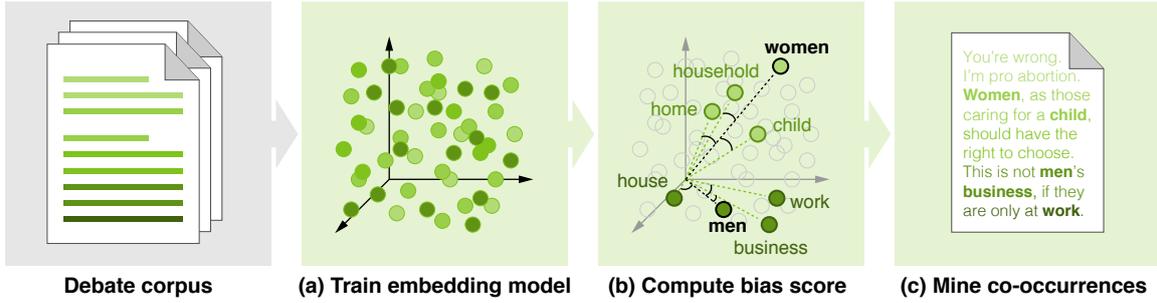
\hwfigure{process}{Overview of the methodology of our experiments: Given a debate corpus, (a) a custom embed\-ding model is trained. (b) A bias score is computed using one of the  \weat\ tests. (c) Word co-occurrences that give hints about the bias are mined from the corpus (the shown example is made-up for emphasis).}

\subsection{Training of Word Embedding Models and Computation of Bias Scores}

The Implicit Association Test~\cite{greenwald_measuring_1998} measures the response times of study participants for pairing concepts based on word lists and uses it as a proxy for bias~\cite{caliskan_semantics_2017}. The test requires four lists of words: two target word lists, $A$~and~$B$, and two association word lists, $X$~and~$Y$. Target word lists implicitly describe a concept, such as a social group, while association word lists describe an association, such as being pleasant or unpleasant. The \weat\ metric~\cite{caliskan_semantics_2017} aims to adapt this method to word embedding models. Under the assumption that word vectors with similar meaning are closer to each other, it computes the mean cosine distance between the four lists in a given model. The score represents the effect size of the difference in distances, which is formulated as follows:
\[
    \frac
        {\text{mean}_{x\in X} s(x, A, B) \;-\; \text{mean}_{y\in Y} s(y, A, B)}
        {\text{std\_dev}_{w\in X\cup Y} s(w, A, B)}
\]
where $s(x, A, B)$ is the difference in cosine distances for each word in $A$ to $x$ and each word in $B$ to $x$. The effect size then acts as a proxy to quantify bias.

\paragraph{Portal-specific Models}

To apply the \weat\ metric to the texts of debate portals, we first extract all posts from the corpora and train one separate custom word embedding model on each corpus using the GloVe algorithm~\cite{pennington_glove_2014}.
Given the custom models, we then evaluate social bias in them, focusing on three of the most common bias types: towards \textit{ethnicity}, \textit{gender}, and \textit{age}~\cite{fiske_social_2004}. These types are roughly represented by seven of the originally proposed \weat\ tests, found in Table \ref{tab:weat-test-overview}. As a notion of stability of the results, we additionally create five embedding models that are trained on random splits of the evaluated corpora and calculate the standard deviation of their \weat\ scores.

\begin{table*}[t]
\small
\centering
\setlength{\tabcolsep}{2pt}
    \begin{tabular}{@{}ll@{$\;\;\;$}ll@{$\;\;\;$}ll@{}}
        \toprule
        \bf  & \bf  & \multicolumn{2}{c}{\bf Target words} & \multicolumn{2}{c}{\bf Association words}  \\
        \cmidrule(l@{0pt}r@{10pt}){3-4}\cmidrule(l@{0pt}r@{0pt}){5-6}
        \bf Type & \bf Test & \bf Compared concepts	& \bf Examples & \bf Compared concepts	& \bf Examples   \\
        \midrule
        & \gr \weat-1 & \gr Flowers vs.\ insects & \gr rose, spider & \gr Pleasant vs.\ unpleasant & \gr freedom, hatred \\
        & \gr \weat-2 & \gr Instruments vs.\ weapons &  \gr guitar, gun & \gr Pleasant vs.\ unpleasant & \gr freedom, hatred \\
        \addlinespace
        Ethnicity & \weat-3 & European- vs.\ African-Amer.\ names & Sara, Alonzo & Pleasant vs.\ unpleasant & freedom, hatred \\
                  & \weat-4 & European- vs.\ African-Amer.\ names & Brad, Darnell & Pleasant vs.\ unpleasant & freedom, hatred \\
                  & \weat-5 & European- vs.\ African-Amer.\ names & Brad, Darnell & Pleasant vs.\ unpleasant &  joy, agony \\
        \addlinespace
        Gender    & \weat-6 & Male vs.\ female names & John, Amy & Career vs.\ family & executive, children \\
                  & \weat-7 & Math vs.\ arts & algebra, poetry & Male vs.\ female terms & man, woman \\
                  & \weat-8 & Science vs.\ arts & physics, symphony & Male vs.\ female terms & father, mother \\
        \addlinespace
                  & \gr \weat-9 & \gr Mental vs.\ physical disease & \gr sad, cancer & \gr  Temporary vs.\ permanent & \gr occasional, chronic \\
        \addlinespace
        Age       & \weat-10 & Young vs.\ old people's names & Tiffany, Bernice & Pleasant vs.\ unpleasant & joy, agony\\
        \bottomrule
    \end{tabular}
    \caption{Overview of the 10 \weat{} tests from~\newcite{caliskan_semantics_2017}. As we focus on \textit{ethnicity}, \textit{gender}, and \textit{age} type, we evaluate the given embedding models only on the tests printed in black. The given examples are drawn from the lists; equal words across tests indicate that the same list was used.}
    \label{tab:weat-test-overview}
\end{table*}

\paragraph{Baseline Models}

To be able to assess the \weat\ results obtained for the trained custom embedding models, we also evaluate two pre-trained models that have shown different levels of bias in previous work and use those in the sense of ``baselines''. They allow us to interpret the results in context. As our upper bias boundary, we choose the GloVe model pre-trained on CommonCrawl data~\cite{pennington_glove_2014}, as it has shown to comprise a high level of bias of different types~\cite{caliskan_semantics_2017,sweeney_transparent_2019}. Similarly, we use the pre-trained Numberbatch model 19.08~\cite{speer_2017a} as the lower boundary. Not only is this model claimed to be debiased in multiple ways~\cite{speer_2017b}, it has also been shown to be the least biased compared to other pre-trained models~\cite{sweeney_transparent_2019}.

\paragraph{Group-specific Models}

As indicated in Section~\ref{sec:data}, the debate.org corpus comes with detailed meta-information about several users. To gain insights into the bias of different user groups, we therefore repeat the process outlined above for posts from self-identified ``black'' and ``white'' users,%
\footnote{In principle, we refrain from using those terms to refer to ethnicities, as they reduce groups to skin color, are thus stereotypes and not adequate to represent the groups~\cite{bryc_genetic_2015}, promoting topological thinking~\cite{jorde_genetic_2004}. However, as they are present on debate.org, we use them here to refer to the respective user groups that self-identified as ``black'' or ``white''.}
female and male users, as well as users below the age of 23 and of 23+%
\footnote{The age of a user is the number of years from the specified birthday to 2017, the year of the data collection~\cite{durmus_corpus_2019}. In all our tests, we assumed those information to be true. As we identified some anomalies, e.g. more than 200 users are assigned an unlikely age of 118 years, we acknowledge that the results might not be particularly reliable.}
in the debate.org dataset. While the age threshold may seem random, age data was sparse and, so, the boundaries were chosen to maximize balance, in order to allow for a rough evaluation of ``younger'' and ``older'' users. More or less, the three chosen pairs of user groups coincide with the evaluated social groups and, thus, may provide insightful comparisons.

\subsection{Mining and Analysis of Word Co-occurrences}

We further conduct a preliminary word co-occurrence analysis for each corpus with the aim to gain additional insights into the observed \weat\ results. To achieve this, we first remove noise, such as stopwords, punctuation, and URLs from the corpora. Afterwards, we take as input all social group identity words of the \weat\ tests, as exemplified in Table \ref{tab:weat-test-overview}. For each list, we extract all words co-occurring with the words in the lists in a window size of 20 (ten words to the left and ten words to the right of the target word). Next, we count their total occurrences and manually evaluate the 100 most common words. As a by-product, we also retrieve the total number of occurrences of target words, which further contribute towards an understanding of the \weat\ results.

As the \weat\ evaluations only analyze the word embedding models, there is no notion of how often two target and association words are actually used in nearby context. Thus, to better understand such relations, we additionally evaluate the co-occurrences of words in the target and association lists on the sentence level. For each \weat\ test, we first build all possible word pairs from the four lists in order to then filter the posts. If a post does not contain both words in a pair, it is discarded. The remaining posts are then split into sentences, allowing us to finally count the co-occurrences of each pair.
    \section{Results}
\label{sec:results}

\paragraph{Portal-specific and Baseline Models}

The \weat\ results on the embedding models of the complete corpora in comparison to the baseline models can be seen in Table~\ref{tab:results-weat-portals}. Except for the \weat-8 test, GloVe CommonCrawl yields the highest bias values (ranging from 1.0896 to 1.8734) and Numberbatch yields rather low bias values, respectively. Among the corpora, the Internet Argument Corpus v2~(IAC) shows the lowest values on average (e.g., 0.1300 for \weat-5) and thus seems to be the least biased. In contrast, the debate.org corpus has higher values in almost all tests, with a noteworthy difference in the gender bias tests \weat-6 and \weat-8. In cases where it does not have the highest values (as for \weat-6), it is surpassed by the CMV corpus. Compared to the \weat\ scores of the baseline models, the results of all debate corpora are mostly closer to the debiased Numberbatch model than to GloVe CommonCrawl.

Regarding the general direction of the \weat\ scores, we see in Table~\ref{tab:results-weat-portals} that most of the observed effect sizes are positive, indicating a closer association of the first list of target words with the first list of association words (see Table~\ref{tab:weat-test-overview} for the list ordering). This means that (a)~European-American names are more associated with pleasant terms than African-American ones, (b)~male names/terms are more closely associated with career, math, and science terms than female ones, and (c)~young people's names are more associated with pleasant words that old people's names.

\begin{table*}[t]
\small
\centering
\setlength{\tabcolsep}{4pt}
    \begin{tabular}{llrrrrrrr}
        \toprule
        & & \multicolumn{3}{c}{\textbf{(a) Ethnicity}} & \multicolumn{3}{c}{\textbf{(b) Gender}} & \multicolumn{1}{c}{\textbf{(c) Age}} \\
        \cmidrule(lr){3-5} \cmidrule(lr){6-8} \cmidrule(l){9-9}
        \bf Type & \bf Embedding Model & \bf \weat-3 & \bf \weat-4 & \bf \weat-5 & \bf \weat-6 & \bf \weat-7 & \bf \weat-8 & \bf \weat-10 \\
        \midrule
        Pretrained	& Numberbatch (debiased) & 0.3203 & \em --0.0994 & 0.5621 & 1.7527 & \em 0.0153 & 0.7429 & 0.8023 \\
        & GloVe CommonCrawl & \bf 1.4367 & \bf 1.5778 & \bf1.3803 & \bf 1.8734 & \bf 1.0896 & 1.2780 & \bf 1.2527 \\
        \addlinespace
        Custom & IAC & 0.2933 & 0.3624 & \em 0.1300 & \em--0.3396 & 0.4009 & 0.6265 & \em 0.3208 \\
        & CMV & \em --0.0632 & 0.4956 & 0.4175 & 1.3151 & --0.3055 & \em 0.4626 & 0.7839 \\
        & debate.org: Full corpus & 0.4125 & 0.5742 & 0.5811 & 1.2775 & 0.5833 & \bf 1.3053 & 0.4018 \\
        \bottomrule
    \end{tabular}
    \caption{Bias values of the custom embedding models, trained on the three debate portal corpora, in comparison to the pretrained baseline models according to the  \weat\ metric. The higher the absolute value, the larger the bias. The highest value in each column is marked bold, the lowest italicized.}
    \label{tab:results-weat-portals}
\end{table*}

\begin{table*}[t]
\small
\centering
\setlength{\tabcolsep}{4pt}
    \begin{tabular}{lrrrrrrr}
        \toprule
        & \multicolumn{3}{c}{\textbf{(a) Ethnicity}} & \multicolumn{3}{c}{\textbf{(b) Gender}} & \multicolumn{1}{c}{\textbf{(c) Age}} \\
        \cmidrule(lr){2-4} \cmidrule(lr){5-7} \cmidrule(l){8-8}
        \bf Embedding Model & \bf \weat-3 & \bf \weat-4 & \bf \weat-5 & \bf \weat-6 & \bf \weat-7 & \bf \weat-8 & \bf \weat-10 \\
        \midrule
        debate.org: ethnicity-black & 0.2483 & \bf 1.1304 & \bf 1.5238 & --0.2073 & --0.0007 & 0.7710 & n/a \\
        debate.org: ethnicity-white & --0.1915 & --0.4062 & \em 0.2535 & 0.5080 & \em 0.0006 & \bf 1.3182 & 0.4592 \\
        \addlinespace
        debate.org: gender-female & 0.3198 & 0.8689 & 1.3645 & 0.1251 & \bf 1.0460 & 0.7991 & 0.9661 \\
        debate.org: gender-male & \em 0.0363 & \em 0.2730 & 0.8751 & \bf 0.7665 & 0.8377 & 1.3231 & \em 0.2112 \\
        \addlinespace
        debate.org: age-below-23 & \bf --1.7250 & 0.4406 & --1.1198 & --0.2220 & 0.5045 & \em 0.5158 & n/a \\
        debate.org: age-23-up & --0.9210 & 0.8369 & 0.5143 & \em 0.0624 & 0.1848 & 0.6525 & \bf 2.0051 \\
        \bottomrule
    \end{tabular}
    \caption{Bias values for the group-specific embedding models of the debate.org corpus, according~to~\weat. Higher absolute values mean larger bias, the highest  in each column is marked bold, the lowest italicized. \weat-10 has no result for {\em ethnicity-black} and {\em age-below-23} due to too many out-of-vocabulary tokens. The \weat-10 value above $2$ of the {\em age-23-up} corpus is probably caused by a floating point error.}
    \label{tab:results-weat-ddo}
\end{table*}

\paragraph{Group-specific Models}

A similar observation can be made for the debate.org sub-corpora in Table~\ref{tab:results-weat-ddo}. Especially for the embedding models based on posts of female and black users, though, the standard deviation of the \weat\ bias values is higher than for the whole corpus, suggesting less reliable results. For the respective user groups, the \weat\ scores further seem to indicate a bias against the own social group. For example, the \weat-3 and \weat-4 test of the models for black and white users indicate closer associations to pleasant terms with the respective other social group (positive values for black, negative values for white). For female users, the analog notion applies to \weat-6, \weat-7 and \weat-8. Another interesting observation can be made for the age groups: While the ethnicity \weat\ tests indicate that older users are more biased towards European-American names, the exact opposite is true for younger users. In general, depending on the specific \weat\ test, posts from all user groups seem to be biased in different regards. For example, while on average female users have the highest \weat\ values, male users seem to be slightly more biased towards genders. Similarly, posts of younger users show the highest \weat\ values in the ethnicity tests.

With some exceptions, the overall direction of the \weat\ results in Tables~\ref{tab:results-weat-portals} and~\ref{tab:results-weat-ddo} suggests that the three evaluated debate corpora are all biased towards men (compared to women) and the European-American ethnicity group (compared to the African-American group). That said, when building future \gls{ca} systems, the IAC corpus is probably the best choice to reduce social biases in general. It is important to note, though, that it is also the smallest corpus of the three and does not include user information. The CMV corpus seems like a good compromise between the two, as it received lower \weat\ values than the debate.org corpus and offers more data at the same time.

\paragraph{Word Co-occurrences}

The co-occurrence counts in Table~\ref{tab:results-weat-cooccurrences} confirm the observed bias values in some cases. In the debate.org corpus, male identity words indeed occur more often with the math-related terms of the \weat-7 test compared to female identity words. As a specific example, the word ``he'' co-occurs with math-related terms $1662$ times, while the female counterpart ``she'' is only mentioned $178$ times in a sentence with the same class of words. Similarly, in the CMV corpus, science-related terms of the \weat-8 test appear $2006$ times in the same sentence as the male identity word ``his'' while only sharing the same sentence $1396$ times with all female identity words {\em combined}.

In most cases, however, words from both groups are only rarely mentioned in the same sentences, compared to the overall size of the corpora. For the tests that use names as social group identifiers, e.g. \weat-6, this problem is even more noticeable, as indicated by the lower numbers presented in Table~\ref{tab:results-weat-cooccurrences}. In all three corpora, names generally co-occur less often with the association words in the same sentence compared to terms that describe a concept, as used, for example, in \weat-7 and \weat-8. This not only makes it harder to directly interpret the \weat\ results. It also indicates that the cosine distance, on which the \weat\ score is based, relies more on either a distant context, such as an entire post, or on common co-occurrences with potentially unrelated words. That said, the results presented above should be interpreted with care.
    \begin{table*}[t]
\centering
\setlength{\tabcolsep}{4.25pt}
\small
    \begin{tabular}{lcccccc}
        \toprule
        \bf (a) Ethnicity&  \multicolumn{2}{c}{\bf \weat-3}  &  \multicolumn{2}{c}{\bf \weat-4}  &  \multicolumn{2}{c}{\bf \weat-5}  \\
        \cmidrule(lr){2-3}\cmidrule(lr){4-5}\cmidrule(lr){6-7}
        \bf Corpus & \bf European & \bf African & \bf European & \bf African & \bf European & \bf African  \\
        \midrule
        IAC        & 1075:883\phantom{0} & \phantom{0}5:14 & 280:206 & 14:22 & \phantom{0}93:69\phantom{0} & \phantom{0}8:3\phantom{0}  \\
        CMV        & 1503:1461 & 52:35 & 324:257 & 47:26 & 141:80 & 18:19  \\
        debate.org & 1960:2009 & 23:26 & 530:429 & 45:73 & 180:108 & 32:19  \\
        \bottomrule
    \end{tabular}
\\[2ex]
        \begin{tabular}{lcccccc}
        \toprule
        \bf (b) Gender &  \multicolumn{2}{c}{\bf \weat-6}  &  \multicolumn{2}{c}{\bf \weat-7}  &  \multicolumn{2}{c}{\bf \weat-8}  \\
        \cmidrule(lr){2-3}\cmidrule(lr){4-5}\cmidrule(lr){6-7}
        \bf Corpus & \bf Male & \bf Female & \bf Male & \bf Female & \bf Male & \bf Female \\
        \midrule
        IAC        & 313:712\phantom{0} & \phantom{0}35:74\phantom{0} & 1108:732\phantom{0} & \phantom{0}244:156\phantom{0} & 2648:682\phantom{0} & 278:133  \\
        CMV        & 865:1229 & \phantom{0}37:104 & 4586:3976 & 2006:1653 & 5840:4008 & 1396:1255 \\
        debate.org & 456:844\phantom{0} & \phantom{0}31:101 & 3790:1957 & \phantom{0}572:438\phantom{0} & 5351:2011 & 446:405 \\
        \bottomrule
    \end{tabular}
    \quad
  \begin{tabular}{lcc}
        \toprule
        \bf (c) Age &  \multicolumn{2}{c}{\bf \weat-10}\\
        \cmidrule(lr){2-3}
        \bf Corpus & \bf Young & \bf Old \\
        \midrule
        IAC        & 36:19 & 1:2 \\
        CMV        & 73:42 & 5:9 \\
        debate.org & 41:21 & 3:3 \\
        \bottomrule
    \end{tabular}
    \caption{Absolute co-occurrences per \weat\ evaluation between social group identity words and association words. The number left to the colon denotes the count of words from the first association list (e.g. {\em pleasant}), the number right to the colon the count for the second association list (e.g., {\em unpleasant}). Note that for \weat-7 and \weat-8, the numbers denote the counts of the respective target word lists.}
    \label{tab:results-weat-cooccurrences}
\end{table*}

\section{Discussion}

To put the results into context, we will discuss two additional observations more closely in the following, namely the low number of occurrences of some identity words as well as the influence of using names for social group lexicons. We will then close this section with limitations observed during the analysis.

\subsection{Low Occurrence of Identity Words}

In some experiments, the number of occurrences of identity words should be considered when interpreting the results: In the \weat-10 tests, the old people's names used for the social group of elderly people generally have a very low frequency. In the posts of black debate.org users, they sum up to only 18 occurrences in total.  The same is true for users above the age of 22. Even though the overall number of occurrences seems sufficient to conduct the \weat\ evaluation, this definitely calls the meaningfulness of the embedding model into question and, with that, the association tests for those groups.

Also, the occurrence {\em ratio} of identity words of two social groups being compared with \weat\ deserves discussion. For all evaluated tests, at least twice as many identity words of one group are mentioned as for the other, in the highest case even 101 as many. The identity words of the European-American group utilized by \weat-3, for example, occur on average 54.8~times as often as the African-American ones across the three debate corpora. While this ratio is not as high for the other tests, the general tendency is that European-American names are mentioned more often. The same is true for male names and terms, which are used more frequently than female ones, and young people's names that occur more often than old people's names across all corpora. Adding to this is the fact that the identity words are, compared to corpus sizes, only rarely used together in close context, say, in a sentence. The distance between two words in the vector space thus relies mostly on the co-occurrence with other, unrelated words.

On one hand, these observations make it less trivial to interpret  \weat\ results, as the different occurrences cause unequal probabilities for the two identity word lists to co-occur with the tested association words. On the other hand, they imply that the evaluated corpora are not very diverse and, so, provide imbalanced data with respect to the evaluated social groups. For debate.org, this imbalance can also be seen in the size of certain user groups, resulting in an unequal number of arguments from different groups.\,

Another potential confounding factor is the number of out-of-vocabulary (OOV) words. When the total number of identity words is low, many OOV words may make results even more unreliable. Among others, this is the case for the female user posts of the debate.org corpus. African-American names from the \weat-3 test make up 41~of the 49~OOV words, leaving only 9 of the initial 50 names for the association test. A similar case is \weat-10 for users above 22. From the eight old people's names, only one appears in the sub-corpus. An immediate effect of the missing words is, however, not visible from the results.

One practical consequence of the discussed unequal distribution is that automated systems, trained and evaluated on these corpora, may favor arguments of the majority group. The underlying discrimination against minority groups may lead to unfair behavior in all stages of \gls{ca}, e.g., to a better ranking of arguments from majority groups in quality assessments or to generated arguments that reflect the opinion of the majority group mainly. These examples stress the need for more diverse debate corpora representing different social groups in a more balanced way. Creating such corpora will not be easy, though, since detailed user information is not often available on debate portals and other argument sources.

\subsection{Using Names as Social Groups}

A general issue underlying the results is that a list of names does not describe the concept of a social group to a level required by co-occurrence methods such as \weat. For example, the token ``palin'' often co-occurs with the female list word ``sarah'' used in \weat-6, referring to the politician {\em Sarah Palin}. Consequently, in the debate.org corpus, some other highly ranked terms relate to politics, such as ``president'', ``conservative'', and ``supporter''. While we did not observe this for all three corpora, it demonstrates that using names as identity words can cause public persons to somewhat act as representatives. Texts about them thus also influence the associations of the tests, independent of whether they are part of the group or their behavior is representative of it. This discrepancy leads to associations authors of evaluated texts have with the public persons rather than the social group and may ultimately influence the bias evaluation.

Another issues lies in the overall occurrence of names. Male and female {\em names}, for instance, appear less often in the debate texts than male and female {\em terms}. As already discussed above, a smaller number of occurrences of identity words may make the results more prone to distortions, due to small fluctuations in co-occurrences, and thus less reliable in general.

Together, these issues suggest that names might not be appropriate to represent a social group and to analyze bias for the evaluated corpora. Also, it seems questionable whether they are generally statistically representative of the target social groups. Partially, however, the lists of names also led to expected associations, as is the case for the \weat-3, \weat-4 and \weat-5 test on the CMV corpus. The terms co-occurring with the African-American names suggest that they were at least to a small degree able to capture the associations, since some of the most co-occurring terms included ``African'' and ``black''.%
\footnote{We do not suggest that these terms adequately represent and describe the African-American social group, but simply state that the association between the social group and the terms generally holds.}

\subsection{Limitations}

One limitation of our methodology is that it relies on the chosen embedding model to accurately model distances and associations between words. This makes it less applicable to smaller datasets, let alone single texts. Additionally, the influence of multiple factors on \weat\ results is unclear. For example, choosing an alternative algorithm to generate an embedding model may yield different results, e.g., word2vec~\cite{mikolov_distributed_2013} or FastText~\cite{bojanowski_enriching_2017}. Future work should explore more sophisticated methods to analyze social bias that do not depend on embedding models, as they are a point of uncertainty that might never be fully explainable due to the nature of generating the models.

Further, the results presented in this work are limited to the accuracy of \weat. This dependence is problematic for three main reasons: %
First, it assumes that the calculation done by the test accurately models the associations between the lexicons. While \newcite{caliskan_semantics_2017} show that they are able to reproduce results from previous psychology studies with humans, it remains open whether the notion generally applies. Second, it expects that the lexicons of the tests are representative of the social groups they ought to model. Especially with a list of names, however, this assumption can lead to unexpected results and might not hold, as shown above. The same is true for the co-occurrence analysis, which is also based on (and limited to) the social group lexicons of \newcite{caliskan_semantics_2017}. Lastly, the test assumes that all evaluated biases are quantifiable. While this allows for automation, this assumption is certainly questionable, among other reasons because every person might perceive bias differently.

Given those limitations, we would like to emphasize that the results presented in this work are meant as a first evaluation that needs to be further backed up and investigated with more tests in the future.

    \section{Conclusion}
\label{sec:conclusion}

In this paper, we have analyzed social bias in three debate corpora commonly used in \gls{ca} research. To this end, we have trained custom word embedding models and evaluated their associations for multiple social groups using \weat. Further, we have analyzed co-occurrences of terms used to define the social groups. We have found all three corpora to show social bias, mostly towards women and African-American people. According to our evaluation, the smallest corpus, IAC~\cite{abbott_internet_2016}, carries the least social bias, whereas the debate.org corpus \cite{durmus_corpus_2019} shows mostly the highest. The CMV corpus~\cite{al_khatib_exploiting_2020} seems like a middle ground, as it contains the most data and is less biased than the debate.org corpus. In all three corpora, we have found imbalances regarding the representation of the evaluated social groups. 

Future work should investigate additional ways in which social bias may be present in \gls{ca} methods, as the underlying data is not the only source. Other possible causes may be the features selected for a trained model or the way in which a model is applied to a real world problem. We believe that future \gls{ca} corpora should be evaluated and mitigated for social bias. As data is underlying most \gls{ca} methods, it is essential that it is as representative for as many social groups as possible in a balanced manner.

    \bibliographystyle{coling}
    \bibliography{argmining20-debate-portal-bias-lit}

\begin{thebibliography}{}

\bibitem[\protect\citename{Abbott \bgroup et al.\egroup
  }2016]{abbott_internet_2016}
Rob Abbott, Brian Ecker, Pranav Anand, and Marilyn Walker.
\newblock 2016.
\newblock Internet {Argument} {Corpus} 2.0: {An} {SQL} schema for {Dialogic}
  {Social} {Media} and the {Corpora} to go with it.
\newblock In {\em Proceedings of the {Tenth} {International} {Conference} on
  {Language} {Resources} and {Evaluation} ({LREC} 2016)}, pages 4445--4452,
  Portorož, Slovenia. European Language Resources Association (ELRA).

\bibitem[\protect\citename{Ajjour \bgroup et al.\egroup }2019]{ajjour:2019a}
Yamen Ajjour, Henning Wachsmuth, Johannes Kiesel, Martin Potthast, Matthias
  Hagen, and Benno Stein.
\newblock 2019.
\newblock Data acquisition for argument search: The args.me corpus.
\newblock In {\em {KI} 2019: Advances in Artificial Intelligence - 42nd German
  Conference on AI, Kassel, Germany, September 23-26, 2019, Proceedings}, pages
  48--59.

\bibitem[\protect\citename{Al~Khatib \bgroup et al.\egroup
  }2020]{al_khatib_exploiting_2020}
Khalid Al~Khatib, Michael Völske, Shahbaz Syed, Nikolay Kolyada, and Benno
  Stein.
\newblock 2020.
\newblock Exploiting {Personal} {Characteristics} of {Debaters} for
  {Predicting} {Persuasiveness}.
\newblock In {\em Proceedings of the 58th {Annual} {Meeting} of the
  {Association} for {Computational} {Linguistics}}, pages 7067--7072, Online.
  Association for Computational Linguistics.

\bibitem[\protect\citename{Bojanowski \bgroup et al.\egroup
  }2017]{bojanowski_enriching_2017}
Piotr Bojanowski, Edouard Grave, Armand Joulin, and Tomas Mikolov.
\newblock 2017.
\newblock Enriching {Word} {Vectors} with {Subword} {Information}.
\newblock {\em Transactions of the Association for Computational Linguistics},
  5:135--146.
\newblock Publisher: MIT Press.

\bibitem[\protect\citename{Bolukbasi \bgroup et al.\egroup
  }2016]{bolukbasi_man_2016}
Tolga Bolukbasi, Kai-Wei Chang, James~Y Zou, Venkatesh Saligrama, and Adam~T
  Kalai.
\newblock 2016.
\newblock Man is to {Computer} {Programmer} as {Woman} is to {Homemaker}?
  {Debiasing} {Word} {Embeddings}.
\newblock In {\em Advances in {Neural} {Information} {Processing} {Systems}
  29}, pages 4349--4357. Curran Associates, Inc.

\bibitem[\protect\citename{Brown \bgroup et al.\egroup
  }2020]{brown_language_2020}
Tom~B. Brown, Benjamin Mann, Nick Ryder, Melanie Subbiah, Jared Kaplan,
  Prafulla Dhariwal, Arvind Neelakantan, Pranav Shyam, Girish Sastry, Amanda
  Askell, Sandhini Agarwal, Ariel Herbert-Voss, Gretchen Krueger, Tom Henighan,
  Rewon Child, Aditya Ramesh, Daniel~M. Ziegler, Jeffrey Wu, Clemens Winter,
  Christopher Hesse, Mark Chen, Eric Sigler, Mateusz Litwin, Scott Gray,
  Benjamin Chess, Jack Clark, Christopher Berner, Sam McCandlish, Alec Radford,
  Ilya Sutskever, and Dario Amodei.
\newblock 2020.
\newblock Language {Models} are {Few}-{Shot} {Learners}.
\newblock arXiv:2005.14165, Version: 3.

\bibitem[\protect\citename{Bryc \bgroup et al.\egroup }2015]{bryc_genetic_2015}
Katarzyna Bryc, Eric Y. Durand, J.~Michael Macpherson, David Reich, and
  Joanna L. Mountain.
\newblock 2015.
\newblock The {Genetic} {Ancestry} of {African} {Americans}, {Latinos}, and
  {European} {Americans} across the {United} {States}.
\newblock {\em The American Journal of Human Genetics}, 96(1):37--53.

\bibitem[\protect\citename{Caliskan \bgroup et al.\egroup
  }2017]{caliskan_semantics_2017}
Aylin Caliskan, Joanna~J. Bryson, and Arvind Narayanan.
\newblock 2017.
\newblock Semantics derived automatically from language corpora contain
  human-like biases.
\newblock {\em Science}, 356(6334):183--186.

\bibitem[\protect\citename{Chang \bgroup et al.\egroup }2019]{chang_bias_2019}
Kai-Wei Chang, Vinod Prabhakaran, and Vicente Ordonez.
\newblock 2019.
\newblock Bias and {Fairness} in {Natural} {Language} {Processing}.
\newblock In {\em Proceedings of the 2019 {Conference} on {Empirical} {Methods}
  in {Natural} {Language} {Processing} and the 9th {International} {Joint}
  {Conference} on {Natural} {Language} {Processing} ({EMNLP}-{IJCNLP}):
  {Tutorial} {Abstracts}}, Hong Kong, China. Association for Computational
  Linguistics.

\bibitem[\protect\citename{Chen \bgroup et al.\egroup
  }2018]{chen_learning_2018}
Wei-Fan Chen, Henning Wachsmuth, Khalid Al-Khatib, and Benno Stein.
\newblock 2018.
\newblock Learning to {Flip} the {Bias} of {News} {Headlines}.
\newblock In {\em Proceedings of the 11th {International} {Conference} on
  {Natural} {Language} {Generation}}, pages 79--88, Tilburg University, The
  Netherlands. Association for Computational Linguistics.

\bibitem[\protect\citename{Chen \bgroup et al.\egroup }2020]{chen:2020}
Wei-Fan Chen, Khalid Al~Khatib, Henning Wachsmuth, and Benno Stein.
\newblock 2020.
\newblock Analyzing political bias and unfairness in news articles at different
  levels of granularity.
\newblock In {\em Proceedings of the 4th Workshop on Natural Language
  Processing and Computational Social Science}. To appear.

\bibitem[\protect\citename{Dev and Phillips}2019]{dev_attenuating_2019}
Sunipa Dev and Jeff Phillips.
\newblock 2019.
\newblock Attenuating {Bias} in {Word} vectors.
\newblock In {\em The 22nd {International} {Conference} on {Artificial}
  {Intelligence} and {Statistics}}, pages 879--887.

\bibitem[\protect\citename{Durmus and Cardie}2019]{durmus_corpus_2019}
Esin Durmus and Claire Cardie.
\newblock 2019.
\newblock A {Corpus} for {Modeling} {User} and {Language} {Effects} in
  {Argumentation} on {Online} {Debating}.
\newblock In {\em Proceedings of the 57th {Annual} {Meeting} of the
  {Association} for {Computational} {Linguistics}}, pages 602--607, Florence,
  Italy. Association for Computational Linguistics.

\bibitem[\protect\citename{Ethayarajh \bgroup et al.\egroup
  }2019]{ethayarajh_understanding_2019}
Kawin Ethayarajh, David Duvenaud, and Graeme Hirst.
\newblock 2019.
\newblock Understanding {Undesirable} {Word} {Embedding} {Associations}.
\newblock In {\em Proceedings of the 57th {Annual} {Meeting} of the
  {Association} for {Computational} {Linguistics}}, pages 1696--1705, Florence,
  Italy. Association for Computational Linguistics.

\bibitem[\protect\citename{Fan \bgroup et al.\egroup }2019]{fan:2019}
Lisa Fan, Marshall White, Eva Sharma, Ruisi Su, Prafulla~Kumar Choubey, Ruihong
  Huang, and Lu~Wang.
\newblock 2019.
\newblock In plain sight: {M}edia bias through the lens of factual reporting.
\newblock In {\em Proceedings of the 2019 Conference on Empirical Methods in
  Natural Language Processing and the 9th International Joint Conference on
  Natural Language Processing (EMNLP-IJCNLP)}, pages 6343--6349. Association
  for Computational Linguistics.

\bibitem[\protect\citename{Fiske}1993]{fiske_controlling_1993}
Susan~T. Fiske.
\newblock 1993.
\newblock Controlling other people: {The} impact of power on stereotyping.
\newblock {\em American Psychologist}, 48(6):621--628.

\bibitem[\protect\citename{Fiske}1998]{fiske_stereotyping_1998}
Susan~T. Fiske.
\newblock 1998.
\newblock Stereotyping, prejudice, and discrimination.
\newblock In {\em The handbook of social psychology}, volume 1-2, pages
  357--411. McGraw-Hill, New York, NY, US, 4th edition.

\bibitem[\protect\citename{Fiske}2004]{fiske_social_2004}
Susan~T. Fiske.
\newblock 2004.
\newblock {\em Social {Beings}: {A} {Core} {Motives} {Approach} to {Social}
  {Psychology}}.
\newblock J. Wiley.

\bibitem[\protect\citename{Garg \bgroup et al.\egroup }2018]{garg_word_2018}
Nikhil Garg, Londa Schiebinger, Dan Jurafsky, and James Zou.
\newblock 2018.
\newblock Word embeddings quantify 100 years of gender and ethnic stereotypes.
\newblock {\em Proceedings of the National Academy of Sciences},
  115(16):E3635--E3644.

\bibitem[\protect\citename{Greenwald \bgroup et al.\egroup
  }1998]{greenwald_measuring_1998}
Anthony~G. Greenwald, Debbie~E. McGhee, and Jordan L.~K. Schwartz.
\newblock 1998.
\newblock Measuring individual differences in implicit cognition: {The}
  implicit association test.
\newblock {\em Journal of Personality and Social Psychology}, 74(6):1464--1480.

\bibitem[\protect\citename{Hamborg}2020]{hamborg_media_2020}
Felix Hamborg.
\newblock 2020.
\newblock Media {Bias}, the {Social} {Sciences}, and {NLP}: {Automating}
  {Frame} {Analyses} to {Identify} {Bias} by {Word} {Choice} and {Labeling}.
\newblock In {\em Proceedings of the 58th {Annual} {Meeting} of the
  {Association} for {Computational} {Linguistics}: {Student} {Research}
  {Workshop}}, pages 79--87, Online. Association for Computational Linguistics.

\bibitem[\protect\citename{Jones}2019]{jones_machine_2019}
M.~Tim Jones.
\newblock 2019.
\newblock Machine learning and bias.
\newblock
  \url{https://developer.ibm.com/technologies/machine-learning/articles/machine-learning-and-bias/}.
\newblock Last accessed: 2020-09-07.

\bibitem[\protect\citename{Jorde and Wooding}2004]{jorde_genetic_2004}
Lynn~B. Jorde and Stephen~P. Wooding.
\newblock 2004.
\newblock Genetic variation, classification and 'race'.
\newblock {\em Nature Genetics}, 36(11):S28--S33.

\bibitem[\protect\citename{Kienpointner and
  Kindt}1997]{kienpointner_problem_1997}
Manfred Kienpointner and Walther Kindt.
\newblock 1997.
\newblock On the problem of bias in political argumentation: {An} investigation
  into discussions about political asylum in {Germany} and {Austria}.
\newblock {\em Journal of Pragmatics}, 27(5):555--585.

\bibitem[\protect\citename{Lawrence \bgroup et al.\egroup }2017]{lawrence:2017}
John Lawrence, Mark Snaith, Barbara Konat, Katarzyna Budzynska, and Chris Reed.
\newblock 2017.
\newblock Debating {Technology} for {Dialogical} {Argument}: {Sensemaking},
  {Engagement}, and {Analytics}.
\newblock {\em ACM Transactions on Internet Technology}, 17(3):24:1--24:23.

\bibitem[\protect\citename{Lim \bgroup et al.\egroup
  }2020]{lim_annotating_2020}
Sora Lim, Adam Jatowt, Michael Färber, and Masatoshi Yoshikawa.
\newblock 2020.
\newblock Annotating and {Analyzing} {Biased} {Sentences} in {News} {Articles}
  using {Crowdsourcing}.
\newblock In {\em Proceedings of the 12th {Language} {Resources} and
  {Evaluation} {Conference}}, pages 1478--1484, Marseille, France. European
  Language Resources Association.

\bibitem[\protect\citename{Manzini \bgroup et al.\egroup
  }2019]{manzini_black_2019}
Thomas Manzini, Lim Yao~Chong, Alan~W Black, and Yulia Tsvetkov.
\newblock 2019.
\newblock Black is to {Criminal} as {Caucasian} is to {Police}: {Detecting} and
  {Removing} {Multiclass} {Bias} in {Word} {Embeddings}.
\newblock In {\em Proceedings of the 2019 {Conference} of the {North}
  {American} {Chapter} of the {Association} for {Computational} {Linguistics}:
  {Human} {Language} {Technologies}, {Volume} 1 ({Long} and {Short} {Papers})},
  pages 615--621. Association for Computational Linguistics.

\bibitem[\protect\citename{Mikolov \bgroup et al.\egroup
  }2013]{mikolov_distributed_2013}
Tomas Mikolov, Ilya Sutskever, Kai Chen, Greg~S Corrado, and Jeff Dean.
\newblock 2013.
\newblock Distributed {Representations} of {Words} and {Phrases} and their
  {Compositionality}.
\newblock In {\em Advances in {Neural} {Information} {Processing} {Systems}
  26}, pages 3111--3119. Curran Associates, Inc.

\bibitem[\protect\citename{Papakyriakopoulos \bgroup et al.\egroup
  }2020]{papakyriakopoulos_bias_2020}
Orestis Papakyriakopoulos, Simon Hegelich, Juan Carlos~Medina Serrano, and
  Fabienne Marco.
\newblock 2020.
\newblock Bias in word embeddings.
\newblock In {\em Proceedings of the 2020 {Conference} on {Fairness},
  {Accountability}, and {Transparency}}, {FAT}* '20, pages 446--457, Barcelona,
  Spain. Association for Computing Machinery.

\bibitem[\protect\citename{Pennington \bgroup et al.\egroup
  }2014]{pennington_glove_2014}
Jeffrey Pennington, Richard Socher, and Christopher Manning.
\newblock 2014.
\newblock {GloVe}: {Global} {Vectors} for {Word} {Representation}.
\newblock In {\em Proceedings of the 2014 {Conference} on {Empirical} {Methods}
  in {Natural} {Language} {Processing} ({EMNLP})}, pages 1532--1543, Doha,
  Qatar. Association for Computational Linguistics.

\bibitem[\protect\citename{Rios \bgroup et al.\egroup
  }2020]{rios_quantifying_2020}
Anthony Rios, Reenam Joshi, and Hejin Shin.
\newblock 2020.
\newblock Quantifying 60 {Years} of {Gender} {Bias} in {Biomedical} {Research}
  with {Word} {Embeddings}.
\newblock In {\em Proceedings of the 19th {SIGBioMed} {Workshop} on
  {Biomedical} {Language} {Processing}}, pages 1--13, Online. Association for
  Computational Linguistics.

\bibitem[\protect\citename{Rudinger \bgroup et al.\egroup
  }2018]{rudinger_gender_2018}
Rachel Rudinger, Jason Naradowsky, Brian Leonard, and Benjamin Van~Durme.
\newblock 2018.
\newblock Gender {Bias} in {Coreference} {Resolution}.
\newblock In {\em Proceedings of the 2018 {Conference} of the {North}
  {American} {Chapter} of the {Association} for {Computational} {Linguistics}:
  {Human} {Language} {Technologies}, {Volume} 2 ({Short} {Papers})}, pages
  8--14, New Orleans, Louisiana. Association for Computational Linguistics.

\bibitem[\protect\citename{Sap \bgroup et al.\egroup }2020]{sap_social_2020}
Maarten Sap, Saadia Gabriel, Lianhui Qin, Dan Jurafsky, Noah~A. Smith, and
  Yejin Choi.
\newblock 2020.
\newblock Social {Bias} {Frames}: {Reasoning} about {Social} and {Power}
  {Implications} of {Language}.
\newblock In {\em Proceedings of the 58th {Annual} {Meeting} of the
  {Association} for {Computational} {Linguistics}}, pages 5477--5490, Online.
  Association for Computational Linguistics.

\bibitem[\protect\citename{Speer \bgroup et al.\egroup }2017]{speer_2017a}
Robyn Speer, Joshua Chin, and Catherine Havasi.
\newblock 2017.
\newblock {ConceptNet} 5.5: {An} {Open} {Multilingual} {Graph} of {General}
  {Knowledge}.
\newblock In {\em Thirty-{First} {AAAI} {Conference} on {Artificial}
  {Intelligence}}, San Francisco, California USA.

\bibitem[\protect\citename{Speer}2017]{speer_2017b}
Robyn Speer.
\newblock 2017.
\newblock {ConceptNet} {Numberbatch} 17.04: {B}etter, less-stereotyped word
  vectors.
\newblock
  \url{https://blog.conceptnet.io/posts/2017/conceptnet-numberbatch-17-04-better-}
  \url{less-stereotyped-word-vectors/}.
\newblock Last accessed: 2020-09-03.

\bibitem[\protect\citename{Stab and Gurevych}2016]{stab_recognizing_2016}
Christian Stab and Iryna Gurevych.
\newblock 2016.
\newblock Recognizing the {Absence} of {Opposing} {Arguments} in {Persuasive}
  {Essays}.
\newblock In {\em Proceedings of the {Third} {Workshop} on {Argument} {Mining}
  ({ArgMining2016})}, pages 113--118, Berlin, Germany. Association for
  Computational Linguistics.

\bibitem[\protect\citename{Stede and Schneider}2018]{stede:2018}
Manfred Stede and Jodi Schneider.
\newblock 2018.
\newblock {\em Argumentation Mining}.
\newblock Number~40 in Synthesis Lectures on Human Language Technologies.
  Morgan \& Claypool.

\bibitem[\protect\citename{Sun \bgroup et al.\egroup
  }2019]{sun_mitigating_2019}
Tony Sun, Andrew Gaut, Shirlyn Tang, Yuxin Huang, Mai ElSherief, Jieyu Zhao,
  Diba Mirza, Elizabeth Belding, Kai-Wei Chang, and William~Yang Wang.
\newblock 2019.
\newblock Mitigating {Gender} {Bias} in {Natural} {Language} {Processing}:
  {Literature} {Review}.
\newblock In {\em Proceedings of the 57th {Annual} {Meeting} of the
  {Association} for {Computational} {Linguistics}}, pages 1630--1640, Florence,
  Italy. Association for Computational Linguistics.

\bibitem[\protect\citename{Sweeney and Najafian}2019]{sweeney_transparent_2019}
Chris Sweeney and Maryam Najafian.
\newblock 2019.
\newblock A {Transparent} {Framework} for {Evaluating} {Unintended}
  {Demographic} {Bias} in {Word} {Embeddings}.
\newblock In {\em Proceedings of the 57th {Annual} {Meeting} of the
  {Association} for {Computational} {Linguistics}}, pages 1662--1667, Florence,
  Italy. Association for Computational Linguistics.

\bibitem[\protect\citename{Vanmassenhove \bgroup et al.\egroup
  }2018]{vanmassenhove_getting_2018}
Eva Vanmassenhove, Christian Hardmeier, and Andy Way.
\newblock 2018.
\newblock Getting {Gender} {Right} in {Neural} {Machine} {Translation}.
\newblock In {\em Proceedings of the 2018 {Conference} on {Empirical} {Methods}
  in {Natural} {Language} {Processing}}, pages 3003--3008, Brussels, Belgium.
  Association for Computational Linguistics.

\bibitem[\protect\citename{Wachsmuth \bgroup et al.\egroup
  }2017]{wachsmuth:2017e}
Henning Wachsmuth, Martin Potthast, Khalid Al-Khatib, Yamen Ajjour, Jana
  Puschmann, Jiani Qu, Jonas Dorsch, Viorel Morari, Janek Bevendorff, and Benno
  Stein.
\newblock 2017.
\newblock Building an argument search engine for the web.
\newblock In {\em Proceedings of the 4th Workshop on Argument Mining}, pages
  49--59. Association for Computational Linguistics.

\bibitem[\protect\citename{Zenker}2011]{zenker_experts_2011}
Frank Zenker.
\newblock 2011.
\newblock Experts and {Bias}: {When} is the {Interest}-{Based} {Objection} to
  {Expert} {Argumentation} {Sound}?
\newblock {\em Argumentation}, 25(3):355.

\end{thebibliography}
\end{document}